\documentclass{article}

\usepackage{microtype}
\usepackage{graphicx}
\graphicspath{{figures/}}
\usepackage{floatrow}
\usepackage{booktabs} 
\usepackage[label font=bf,labelformat=parens]{subfig}
\usepackage{hyperref}

\usepackage[utf8]{inputenc}
\usepackage[T1]{fontenc}
\usepackage{amsmath, amsfonts, bm, amssymb, mathtools, commath}
\usepackage{xcolor}
\usepackage{float}
\usepackage{multirow}

\newcommand{\M}{\mathcal{M}}
\renewcommand{\H}{\mathcal{H}}
\renewcommand{\P}{\mathcal{P}}
\renewcommand{\O}{\mathcal{O}}

\usepackage[arxiv]{icml2021} 

\icmltitlerunning{Variational Data Assimilation with a Learned Inverse Observation Operator}

\begin{document}

\twocolumn[
\icmltitle{Variational Data Assimilation with a Learned Inverse Observation Operator}

\icmlsetsymbol{equal}{*}
\begin{icmlauthorlist}
\icmlauthor{Thomas Frerix}{google,tum}
\icmlauthor{Dmitrii Kochkov}{google}
\icmlauthor{Jamie A. Smith}{google}
\icmlauthor{Daniel Cremers}{tum}
\\
\icmlauthor{Michael P. Brenner}{google,harvard}
\icmlauthor{Stephan Hoyer}{google}
\end{icmlauthorlist}

\icmlaffiliation{google}{Google Research}
\icmlaffiliation{harvard}{Harvard University}
\icmlaffiliation{tum}{Technical University of Munich}

\icmlcorrespondingauthor{Thomas Frerix}{thomas.frerix@tum.de}
\icmlcorrespondingauthor{Stephan Hoyer}{shoyer@google.com}

\icmlkeywords{Physics, Machine Learning, Data Assimilation, Non-Convex Optimization}

\vskip 0.3in
]

\printAffiliationsAndNotice{}

\begin{abstract}
Variational data assimilation optimizes for an initial state of a dynamical system such that its evolution fits observational data.
The physical model can subsequently be evolved into the future to make predictions.
This principle is a cornerstone of large scale forecasting applications such as numerical weather prediction.
As such, it is implemented in current operational systems of weather forecasting agencies across the globe.
However, finding a good initial state poses a difficult optimization problem in part due to the non-invertible relationship between physical states and their corresponding observations.
We learn a mapping from observational data to physical states and show how it can be used to improve optimizability.
We employ this mapping in two ways: to better initialize the non-convex optimization problem, and to reformulate the objective function in better behaved physics space instead of observation space.
Our experimental results for the Lorenz96 model and a two-dimensional turbulent fluid flow demonstrate that this procedure significantly improves forecast quality for chaotic systems.
\end{abstract}

\section{Introduction}
\label{sec:introduction}
\begin{figure}[t!]
    \centering
    \sidesubfloat[]{
        \includegraphics[width=0.9\columnwidth]{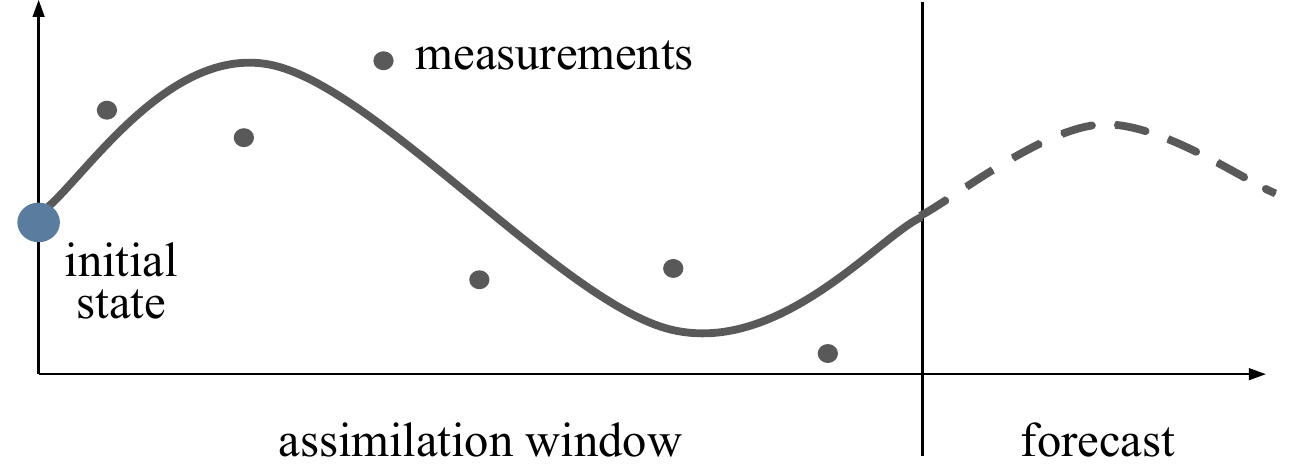}
        \label{fig:variational_da_schema}
    }\\
    \sidesubfloat[]{
        \includegraphics[width=0.9\columnwidth]{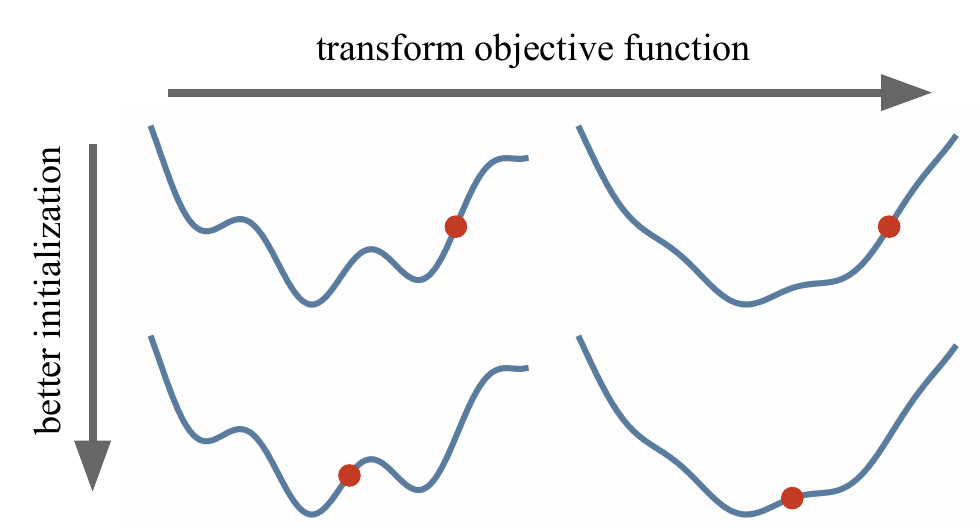}
        \label{fig:improving_opt_schema}
    }\\
    \sidesubfloat[]{
        \includegraphics[width=0.9\columnwidth]{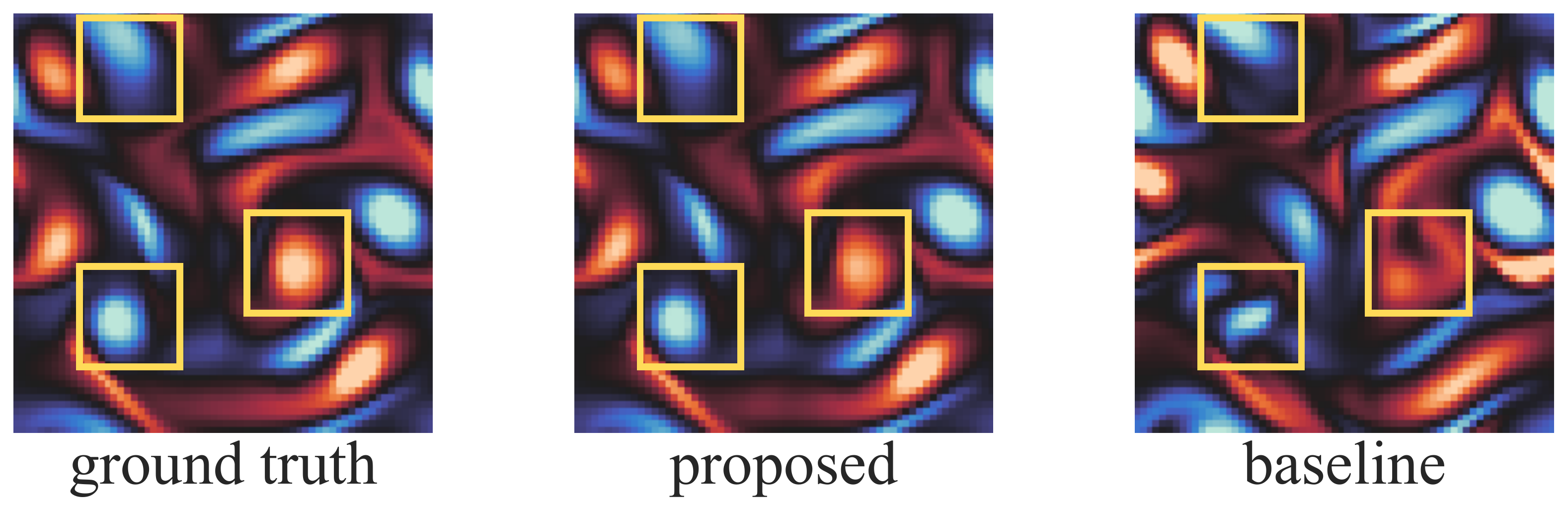}
        \label{fig:result_teaser}
    }\\
    \caption{
        Overview of the proposed method.
        \textbf{(a)} The principle of variational data assimilation. 
        The goal is to optimize for an initial state (large blue dot) of a physical system such that the evolution over an assimilation window (solid line) is close to measurements (small gray dots). 
        The model is subsequently used to make predictions into the future (dashed line).
        \textbf{(b)} Improving optimizability of variational data assimilation.
        We use a learned inverse observation operator to better initialize the optimization problem (red dots) and to transform the non-convex objective function to be better behaved.
        \textbf{(c)} Data assimilation results of the proposed method compared with a traditional algorithm. Depicted is a vorticity prediction of a two-dimensional turbulent fluid flow. 
        The proposed method more accurately captures vorticity features (yellow squares).
    }
\end{figure}

Variational data assimilation provides the basis for numerical weather prediction \cite{ECMWF2019}, integrating the non-linear partial differential equations describing the atmosphere.
The core algorithm is an optimization problem for the initial state of the system, such that when the equations of motion are evolved over time, the resulting trajectories are close to the measurements.
Continuing to evolve the physical system into the future then yields a forecast 
(Figure~\ref{fig:variational_da_schema}).  
Over the last decades, these algorithms have led to a steady improvement in forecast quality, though further improvements are limited by computational resources.
Data assimilation accounts for a significant fraction of the computational cost for numerical weather prediction.
This restricts the amount of data that can be assimilated and only a small volume of available satellite data is utilized for operational forecasts \cite{Bauer2015, Gustafsson2018}.  

Weather forecasting systems are complex algorithmic pipelines \cite{Bauer2015}.
Recent work has shown that in some cases forecasts can be improved by replacing the entire system with a machine learned prediction (see, e.g., \cite{Sonderby2020, Ham2019}).  
This approach is very powerful, but physical models remain more accurate for global weather forecasting \cite{Rasp2020}.
Moreover, they offer guarantees of generalization, interpretability and physical consistency because they are built upon well-known physical principles.
In fact, some of the best pure machine learning approaches rely upon pre-training with simulation data due to insufficient historical observations \cite{Rasp2020b, Ham2019}.
Additionally, physical modeling facilitates the principled coupling of processes on different characteristic spatial and time scales, e.g., the atmosphere, ocean, and land surfaces, which is critical for complex forecasting applications \cite{Bauer2015}.
Consequently, a more promising approach may be to \emph{augment} physical models with machine learning \cite{Watt-Meyer2021-bi, Kochkov2021-sh}.

In this paper, we augment a traditional variational data assimilation algorithm with machine learning. 
We use the equations of motion to evolve the dynamical system, while machine learning is used only to improve the optimization problem for calculating the initial state.
To this end, we learn an approximate inverse to the observation operator.
Using this mapping, we provide an effective initialization scheme for the non-convex optimization problem and transform the objective function for the variational data assimilation problem to be better behaved (Figure~\ref{fig:improving_opt_schema}).
We generate observational data from two model problems, a classical model for data assimilation introduced by Lorenz \cite{Lorenz95}, and a turbulent fluid flow in two spatial dimensions.  
We demonstrate that the algorithm enhanced with machine learning leads to a substantial performance improvements over the baseline. 
Figure~\ref{fig:result_teaser} shows an example of an improved forecast.

\section{Related Work}
\label{sec:related_work}
Data assimilation is a suitable formalism for combining physical modeling with machine learning since large scale applications are characterized by rich physics and large amounts of data.
Both approaches can be viewed in the framework of Bayesian inference \cite{Geer2020}.
Machine learning approaches to modify the physical model for data assimilation include a learned correction to an approximate model \cite{Farchi2020, Brajard2020}, training a machine learning model to completely emulate the physics \cite{Brajard2020b}, and learning a forcing term within the weak-constraint 4D-Var formulation \cite{Bonavita2020}.
In contrast, we use an exact physical model and modify the representation of observations using machine learning.
\citet{Mack2020} formulate variational data assimilation in a latent space derived by training an autoeconder.
The dimensionality reduction allows for significantly faster optimization.
However, this approach loses physical guarantees for decoded states.

Integration of dynamical systems is a central component of data assimilation.
However, simulating high-resolution dynamics quickly becomes computationally intractable.
To ameliorate this issue, several recent works combine traditional numerical solvers with machine learning to obtain high-resolution physics from coarser simulations.
MeshfreeFlowNet \cite{Jiang2020} continuously parameterizes the spatial domain by learning an interpolation function for each grid cell.
\citet{Um2020} incorporate a correction operator directly into the numerical solver and train this function to nudge an inaccurate solution towards a more accurate one. 
The authors of \cite{BarSinai2019, Zhuang2020} learn a discretization scheme for PDEs that better captures the unresolved physics, leading to improvements over ad hoc finite difference discretization methods.
Using a fully differentiable computational fluid solver, \citet{Kochkov2021-sh} demonstrate that with this approach the grid resolution can be reduced by an order of magnitude without sacrificing accuracy.  
Similarly, we use a fully differentiable solver for our model systems and our approach may therefore be complemented by such ideas.

Variational data assimilation requires solving a difficult optimization problem.
Our approach of improving optimizability of this problem with machine learning can be contextualized with other works that employ machine learning to transform a physics constrained optimization problem.
In the context of simulating mechanical materials, \citet{Beatson2020} approximate the inner problem of a bi-level optimization problem by a learned function, thus crucially reducing the the computation cost.
To optimize photonic device designs, \citet{Kudyshev2021} train for a compressed design space with an adversarial autoencoder.
This space is then explored using an evolutionary algorithm.
\citet{Ackmann2020} learn a preconditioner to improve the solution of a linear system arising during the integration of a shallow-water model.
As with our approach, the preconditioned system does not suffer from generalization issues of the machine learning model.
We can guarantee a certain performance level by defaulting to a classical method.
Various works improve optimization problems not with a component learned from training data, but by reparameterizing the objective function with a neural network architecture.
The neural network here acts as an overparameterization with a specific inductive bias, e.g., convolutional neural networks for building hierarchical, multi-scale representations \cite{Hoyer2019, Ulyanov2018} or fully-connected networks for continuous representations \cite{Mildenhall2020}.

\section{Variational Data Assimilation}
\label{sec:variational_da}
The state of the art variational data assimilation algorithm is called 4D-Var~\cite{Bannister2017}.
It minimizes an objective function of the form
\begin{align}
    \label{eq:4dvar}
    J(x_0) =& (x_0 - x^b)^T \mathbf{B}^{-1} (x_0 - x^b) \nonumber \\
           &+ \sum_{t=0}^T(\H(x_t) - y_t)^T \mathbf{R}^{-1} (\H(x_t) - y_t) \nonumber\\
    x_{t+1} =& \M(x_{t}).
\end{align}
The goal is to produce a maximum likelihood estimate of the initial state $x_0$ of a trajectory $(x_1, \dots, x_T)$ that is evolved through a physics model $\M$, given a sequence of observations $(y_1, \dots, y_T)$.
The observation operator $\H$, maps physical states into the space of observations.
As an example, the physics model could be the Navier Stokes equations for evolving a weather system, and the observation operator could measure the state of the atmosphere at discrete weather stations.
The loss $J(x_0)$ models the initial condition and conditional distribution of observations as a multi-variate normal distribution.
The first term incorporates a guess for the initial state $x_0$ (the so-called background state $x^b$), where $\mathbf{B}$  is the background covariance matrix, representing the uncertainty about this assumption, i.e., $x_0 \sim \mathcal{N}(x^b, \mathbf{B})$.
Similarly, the matrix $\mathbf{R}$ models the observation error covariance, i.e., $y_t \sim \mathcal{N}(\H(x_t), \mathbf{R})$.
The simplifying assumption of Gaussian background and observation errors may suffer from model misspecification when applied to real-world data~\cite{Bocquet2010}.
Altogether, this amounts to solving a non-linear least squares problem.
We denote the initialization of this optimization problem by \textit{initial condition} and refer to an \textit{initial state} to describe the first state $x_0$ of a physics trajectory.

4D-Var minimizes objective functions of the form \eqref{eq:4dvar} via gradient based optimization.
To forecast a trajectory $(x_1, \dots, x_{T^\prime})$, the objective \eqref{eq:4dvar} is minimized to estimate $x_0$ over a fixed-length window of observations, the so-called assimilation window, which is shifted in time.
The forecast state from a previous assimilation window becomes the background state for the next assimilation window.
Minimizing \eqref{eq:4dvar} is a difficult optimization problem for various reasons~\cite{Andersson2005}:
First, the physics model $\M$ is in general non-linear or even chaotic, so that small changes in the initial state can lead to large changes in an integrated state.
Secondly, the observation operator $\H$ that reduces information from physics trajectories to observations is usually non-invertible and possibly non-linear.

In what follows, we focus on the difficulty posed by the observation operator $\H$, and learn an approximate inverse to $\H$ that maps the observational data to the space of physical states. 
For simplicity, we focus our analysis on a fixed time horizon without a shifting assimilation window.
Moreover, we neglect prior modeling of the initial state, so that we omit the first term in \eqref{eq:4dvar}. 
Finally, we neglect an explicit observation noise model, i.e., we set $\mathbf{R}$ to be the identity matrix. 
This amounts to studying the following simplified version of the 4D-Var model:
\begin{align}
    \label{eq:da_obs_space}
    J(x_0) = \sum_{t=0}^T ||\H(x_t) - y_t||_2^2, \qquad x_{t+1} = \M(x_t)
\end{align}
The method presented in this paper is not restricted to this setting, but equally applies to the general 4D-Var problem.
However, to study the effect of the observation operator $\H$ on the optimization problem, additional aspects of the problem are not necessary.

\section{Learning an Inverse Observation Operator}
\label{sec:inv_obs_learning}
To be precise, we distinguish the space $\P$ of physical states or \textit{physics space} and the space $\O$ of observations or \textit{observation space}.
The observation operator $\H: \P \rightarrow \O$ maps the physics space $\P$ to the observation space $\O$.
The variational data assimilation objective \eqref{eq:da_obs_space} is formulated in observation space.
However, the non-invertibility (and potential non-linearity) of  $\H$ makes minimizing this objective difficult.
The key idea of this paper is to  parameterize an approximate inverse $h_\theta: \O \rightarrow \P$
and to use machine learning to train the parameters $\theta$.
The training target is to map observations to corresponding physical states, in our notation to obtain $h_\theta(y_t) \approx x_t$.
While in practice there is ample training data from historical observations, in this work we revert to simulations for generating training data.

In order to exploit both spatial and temporal correlations, we construct a fully-convolutional architecture in space and time.
Fully-convolutional architectures are natural for several reasons: they use local filters and therefore enforce the locality of the underlying equations of motion.
Additionally, the number of parameters in a convolutional layer does not increase with input size.
This is vital because typical physics models $\M$ are discretized over large grids.

We implement our models for the approximate inverse in JAX \cite{jax2018github} and use Flax as neural network library, with the Adam optimizer \cite{Kingma2015} and learning rate of $10^{-3}$ for training\footnote{\url{https://github.com/googleinterns/invobs-data-assimilation}}.
We also use JAX to implement differentiable simulators for the physics model $\M$ that arises inside the objective function \eqref{eq:da_obs_space} \cite{Kochkov2021-sh}.
Operational numerical weather prediction models similarly make use of differentiable simulators, where they are known as adjoint models.
All models can be trained and optimized on a single NVIDIA V100 GPU.

\begin{figure}[t!]
    \centering
    \includegraphics[width=\columnwidth]{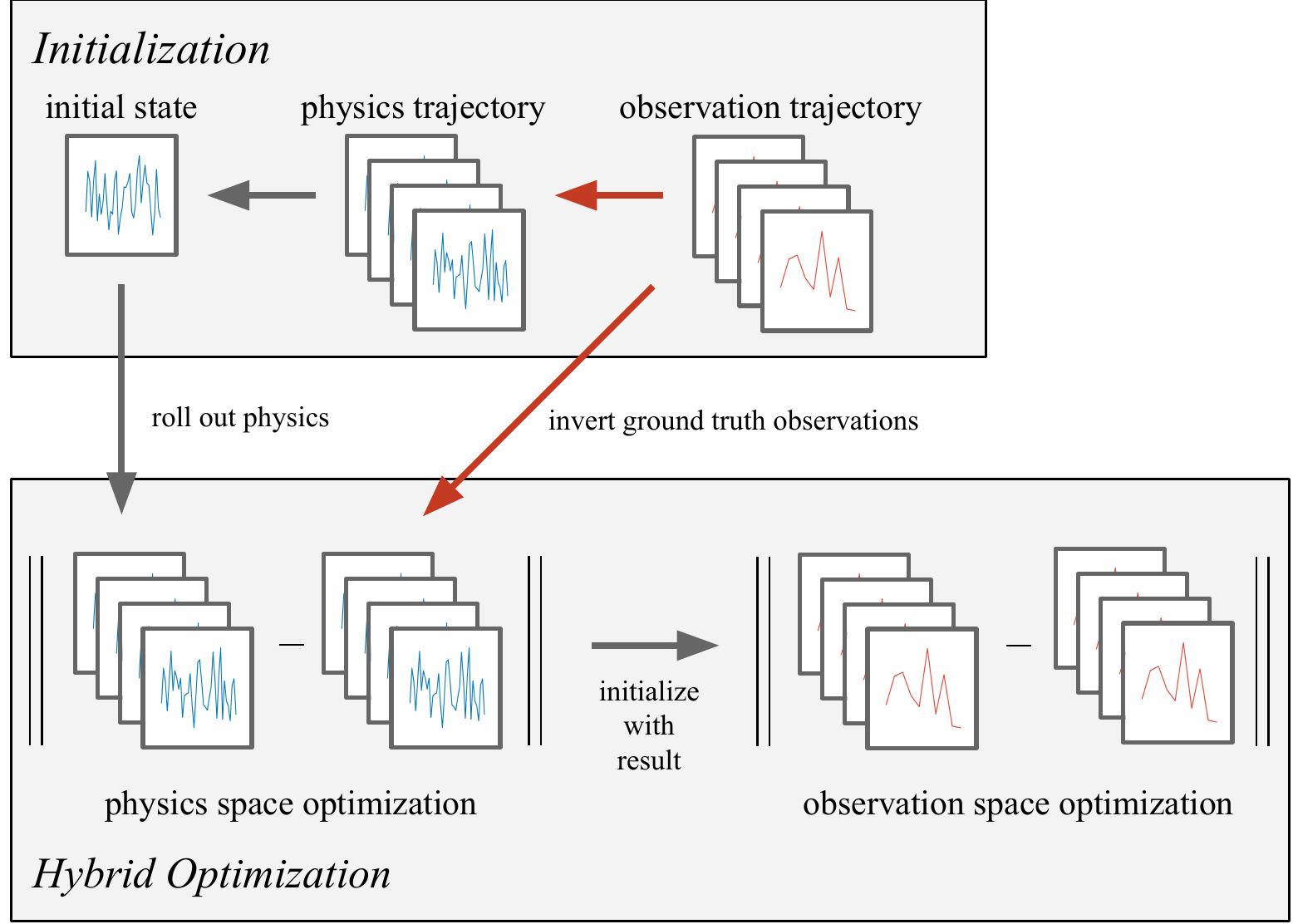}
    \caption{
        Variational data assimilation with a learned inverse observation operator. 
        The learned inverse observation mapping is denoted by red, hollow arrows.
        We approximately invert an observation trajectory and choose its first state as an \textit{initialization} of the non-convex optimization problem.
        The \textit{hybrid optimization} approach first minimizes \eqref{eq:da_physics_space} in physics space and subsequently uses the optimization result to initialize refinement minimization of \eqref{eq:da_obs_space} in observation space.
    }
    \label{fig:pipeline_sketch}
\end{figure}

We use the trained inverse observation operator for two aspects of the optimization problem.
First, we map the earliest trajectory of observations to a trajectory in physics space and then use its first state as an initialization to the optimization problem.
Secondly, we substitute \eqref{eq:da_obs_space} with a reformulated objective function in physics space:
\begin{align}
    \label{eq:da_physics_space}
    \tilde J(x_0) = \sum_{t=0}^T ||x_t - h_\theta(y_t)||_2^2, \qquad x_{t+1} = \M(x_t)
\end{align}
An overview of this method is depicted in Figure~\ref{fig:pipeline_sketch}.
The objective functions \eqref{eq:da_obs_space} and \eqref{eq:da_physics_space} are not equivalent, rather we use \eqref{eq:da_physics_space} as a proxy for \eqref{eq:da_obs_space}.
As we will show in Section~\ref{sec:applications}, minimizing \eqref{eq:da_physics_space} is a more benign optimization problem compared to minimizing \eqref{eq:da_obs_space}.
However, the caveat with minimizing \eqref{eq:da_physics_space} is that we can only expect $h_\theta$ to be an approximation that does not even guarantee to map to a physical state, i.e., a state that one could encounter under the statistically stationary dynamics of the model.
As a consequence, we adopt a \textit{hybrid} approach where we first minimize \eqref{eq:da_physics_space} and use the optimization result to initialize minimizing \eqref{eq:da_obs_space} for further refinement. 

\section{Results}
\label{sec:applications}
We demonstrate this approach on two chaotic dynamical systems, the Lorenz96 model \cite{Lorenz95} and Kolmogorov flow \cite{Chandler2013}.
As our baseline, we follow a common approach \cite{Bannister2008} and assimilate in observation space over a set of uncorrelated variables, i.e., we minimize \eqref{eq:da_obs_space} after a whitening transformation to $\xi = C^{-1/2}x$, where $C$ is the empirical covariance matrix over a set of $10^6$ independent samples from the stationary distribution of the respective dynamical systems.
The empirical covariance matrix might not be positive definite, an issue that is often encountered in applications \cite{Tabeart2020}.
To ensure positive definiteness, we threshold the spectrum of $C$ at $10^{-6}$.
To be precise, we solve
\begin{align}
    \label{eq:da_transform}
    \min_{\xi_0} \sum_{t=0}^T ||\H(C^{1/2}\xi_t) - y_t||_2^2, \; \xi_{t+1} = C^{-1/2}\M(C^{1/2}\xi_t) \;.
\end{align}
We use L-BFGS \cite{Nocedal2006} as an optimizer for assimilation, retaining a history of $10$ vectors for the Hessian approximation.
An optimization step in physics and observation space incurs a comparable computational cost, since the inverse observation operator is applied \textit{prior} to optimization to modify the fitting targets.

We measure the quality of forecasts by an $L_1$  point-wise error metric $\varepsilon$ between two states $z_1, z_2$:
\begin{align}
    \varepsilon(z_1, z_2) \coloneqq  \|z_1 - z_2\|_1 / \gamma\;,
\end{align}
where we scale this metric to a \textit{relative} error by dividing by a mean error $\gamma$ of random independent states sampled from the stationary distribution of the dynamical system.
This metric can be easily interpreted: an order unity error implies the average performance of a random evolution of the system.
We compare the optimized forecasts with the evolution from a ground truth initial state on a set of $100$ test trajectories.

\subsection{Lorenz96 Model}
\label{sec:lorenz96}
The single-level Lorenz96 model \cite{Lorenz95} is a periodic, one-dimensional model where each grid point is evolved according to the equation of motion
\begin{align}
    \frac{d X_k}{dt} = -X_{k-1} (X_{k-2} - X_{k+1}) - X_k + F \;.
\end{align}
Here, the first term models advection, the second term represents a linear damping, and $F$ is an external forcing.
We choose a grid of size $K=40$ and an external forcing $F=8$, parameters where the system is chaotic with a Lyapunov time of approximately $0.6$ time units.
For an observation operator, we use subsampling.
We integrate trajectories over an assimilation window of $T=10$ time steps with a time increment of $\Delta t=0.1$ time units starting from an initial condition in the statistically stationary regime, i.e., where $\sum_k X_k^2$ fluctuates around a constant mean value.

We now demonstrate how a learned inverse observation operator significantly improves forecast results by providing an effective initialization scheme for the non-convex optimization problem and by formulating a more benign objective function in physics space $\P$ instead of observation space $\O$.
As an observation operator for the following experiments, we observe every $4$th grid point.
To approximate the inverse observation operator, we train a fully-convolutional network as described in Table~\ref{tab:lorenz96_net}.
We train on a dataset of $32000$ independent observation trajectories with batch size $8$ for $500$ epochs.
\begin{table}[t]
    \caption{
        Fully-convolutional network for training the inverse observation operator for the Lorenz96 model.
        The table shows a layer with its respective output array dimensions time (\textsc{T}), space (\textsc{X}), and channel (\textsc{C}).
        The \textsc{Conv2D} layer applies periodic convolution in the space dimension and zero-padded convolution in the time dimension.
        The filter size for all convolutional layers is $(3,3)$. 
        \textsc{BN} denotes batch normalization.
        To upscale the grid by a factor of $2$ in layers two and three, we use cubic interpolation.
        As a non-linearity we use the sigmoid-weighted linear unit (\textsc{SiLU}), $\mathrm{silu}(x) = x / (1 + \exp(-x))$.
    }
    \vskip 0.15in
    \centering
    \begin{small}
    \begin{sc}
    \begin{tabular}{ll}
        \toprule
         Layer &  (T, X, C) \\
         \midrule
         Input & (10, 10, 1) \\
         Conv2D + BN + SiLU & (10, 10, 128) \\
         Upsample + Conv2D + BN + SiLU & (10, 20, 64) \\
         Upsample + Conv2D + BN + SiLU & (10, 40, 32) \\
         Conv2D + BN + SiLU & (10, 40, 16) \\
         Conv2D  & (10, 40, 1) \\
         \bottomrule
    \end{tabular}
    \end{sc}
    \end{small}
    \label{tab:lorenz96_net}
\vskip -0.1in
\end{table}

\begin{figure}[t]
    \centering
    \includegraphics[width=0.9\columnwidth]{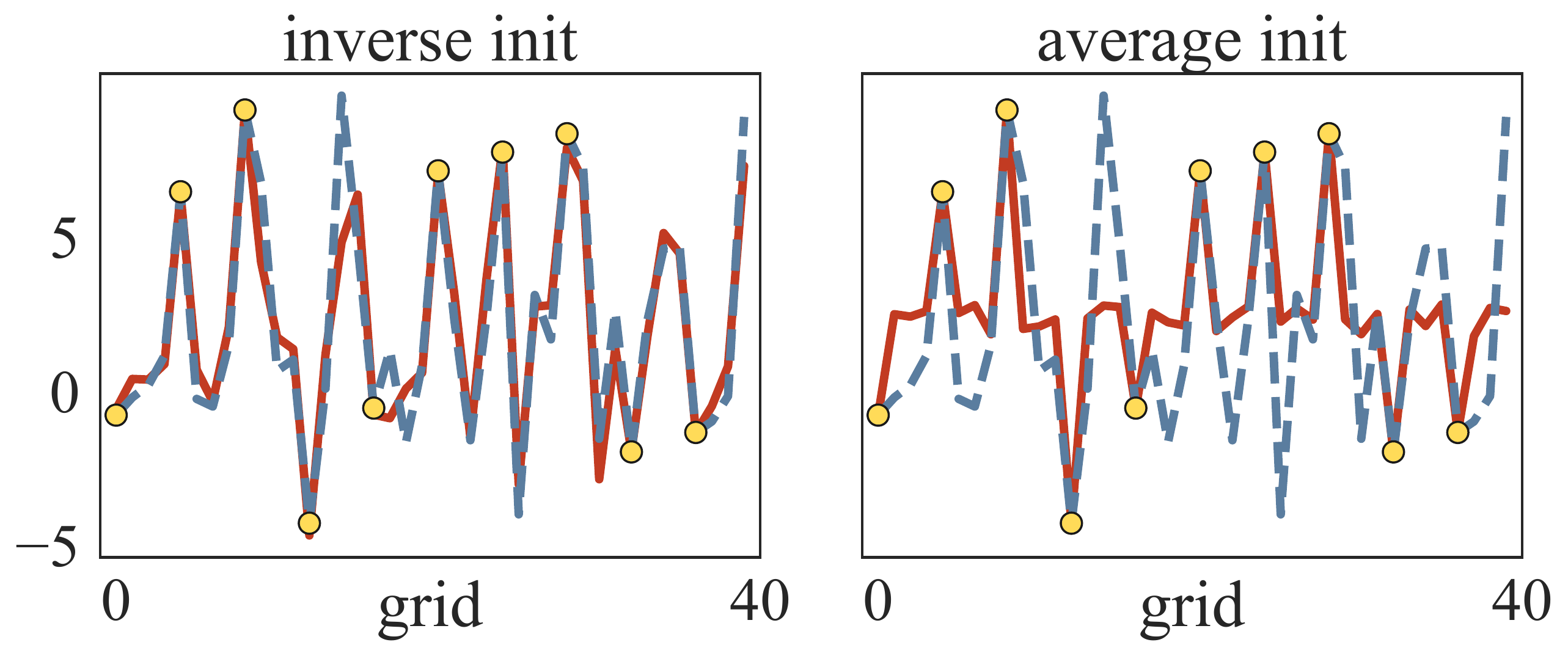}
    \caption{
        Comparison of initialization (solid red) with the ground truth initial state (dashed blue).
        The observed grid points are marked as yellow dots.
        The learned inverse observation mapping takes a trajectory of such subsampled states as input and generates the inverse initialization.
        Inverse initialization is much more accurate than averaging initialization.
    }
    \label{fig:da_init_lorenz96}
\end{figure}

\begin{figure}[t!]
    \centering
    \includegraphics[width=0.95\columnwidth]{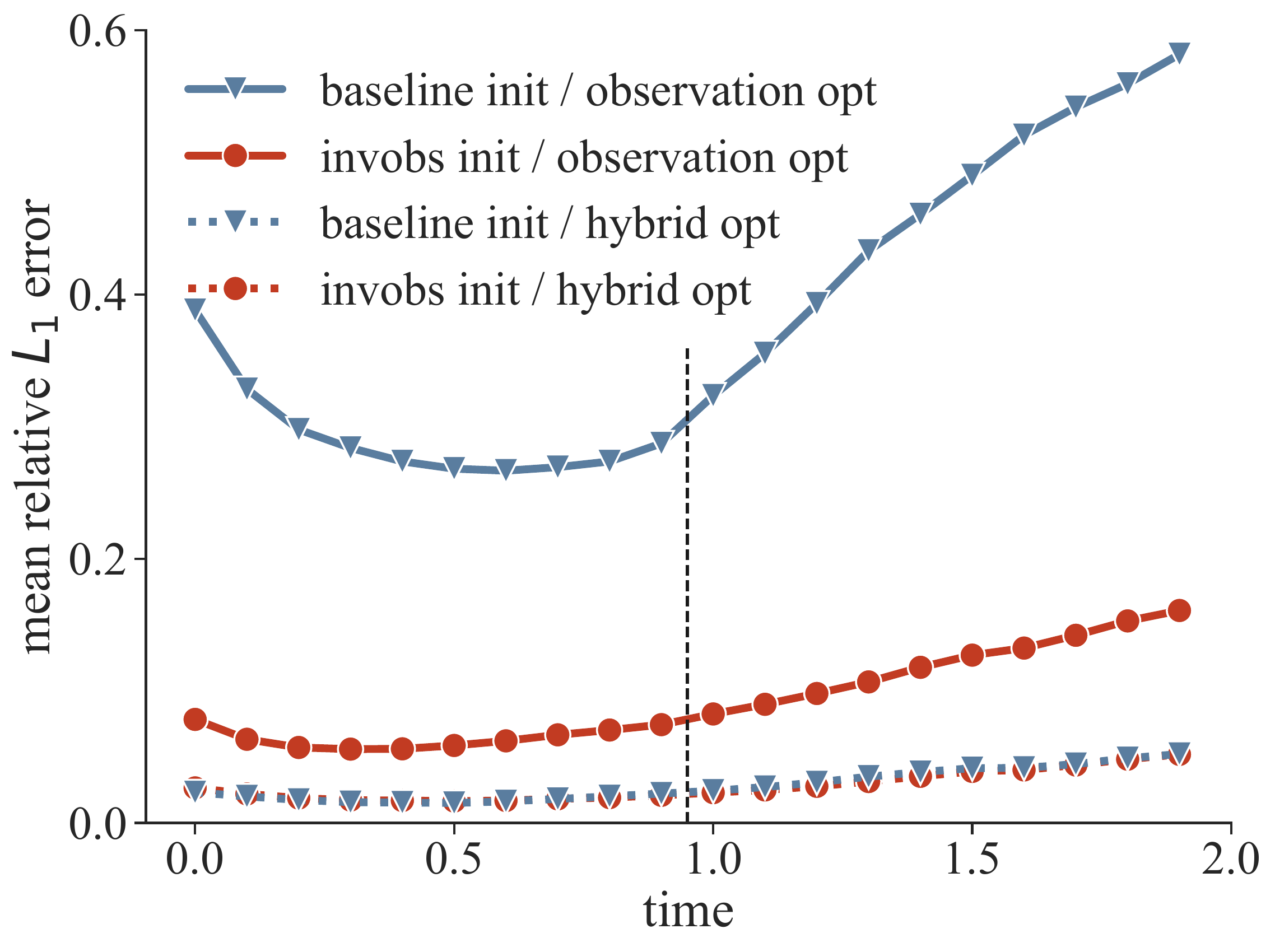}
    \caption{
        Forecast quality with a learned inverse observation operator for the Lorenz96 model. 
        Quality measure is the $L_1$ forecast error relative to a random evolution of the system. Depicted is the mean error based on a sample of $100$ trajectories. 
        The vertical dashed line separates the assimilation window from the forecast window. 
        Inverse initialization improves forecasts for observation space optimization compared with average initialization. 
        For the inaccurate averaging initialization, hybrid optimization significantly improves forecast quality compared with observation space optimization.
        Adding inverse initialization to the hybrid optimization approach leads to a small additional improvement, which is significant with a p-value of $p < 10^{-4}$.
    }
    \label{fig:da_lorenz96_inv_obs}
\end{figure}

For data assimilation, we compare two initialization schemes.
The baseline \textit{averaging initialization} scheme initializes the optimizer with the observed grid points and uses the average over a data set of independent states as an estimate of the unobserved grid points.
This is equivalent to a least-squares fit of the unobserved grid points.
The \textit{inverse initialization} scheme uses the learned inverse observation operator to create the initialization.
To this end, we map a sequence of observations to a physical trajectory and use its first state for initialization as depicted in Figure~\ref{fig:pipeline_sketch}.
Figure~\ref{fig:da_init_lorenz96} shows a qualitative comparison of these two initialization schemes, demonstrating that the learned inverse mapping leads to a much more accurate initialization.
We found that first optimizing for an initial condition from a previous assimilation window, as in 4D-Var, does not improve baseline initialization. 
We compare optimizing in observation space (baseline) with the hybrid approach of first optimizing in physics space and then refining the results in observation space.
For a fair comparison, both optimization methods are limited to $500$ optimization steps with the hybrid method assigning $100$ of these steps to optimization in physics space and the remaining $400$ steps to refinement in observation space. 
The forecast results are shown in Figure~\ref{fig:da_lorenz96_inv_obs}.
Inverse initialization improves forecasts for observation space optimization compared with average initialization.
For the inaccurate averaging initialization, hybrid optimization significantly improves forecast quality compared with observation space optimization.
This suggests that by first optimizing in physics space, we obtain an initialization for refinement in observation space that is located at a favorable basin of attraction.
Adding inverse initialization to the hybrid optimization approach leads to a small additional improvement.

\begin{figure}[t!]
    \centering
    \includegraphics[width=0.75\columnwidth]{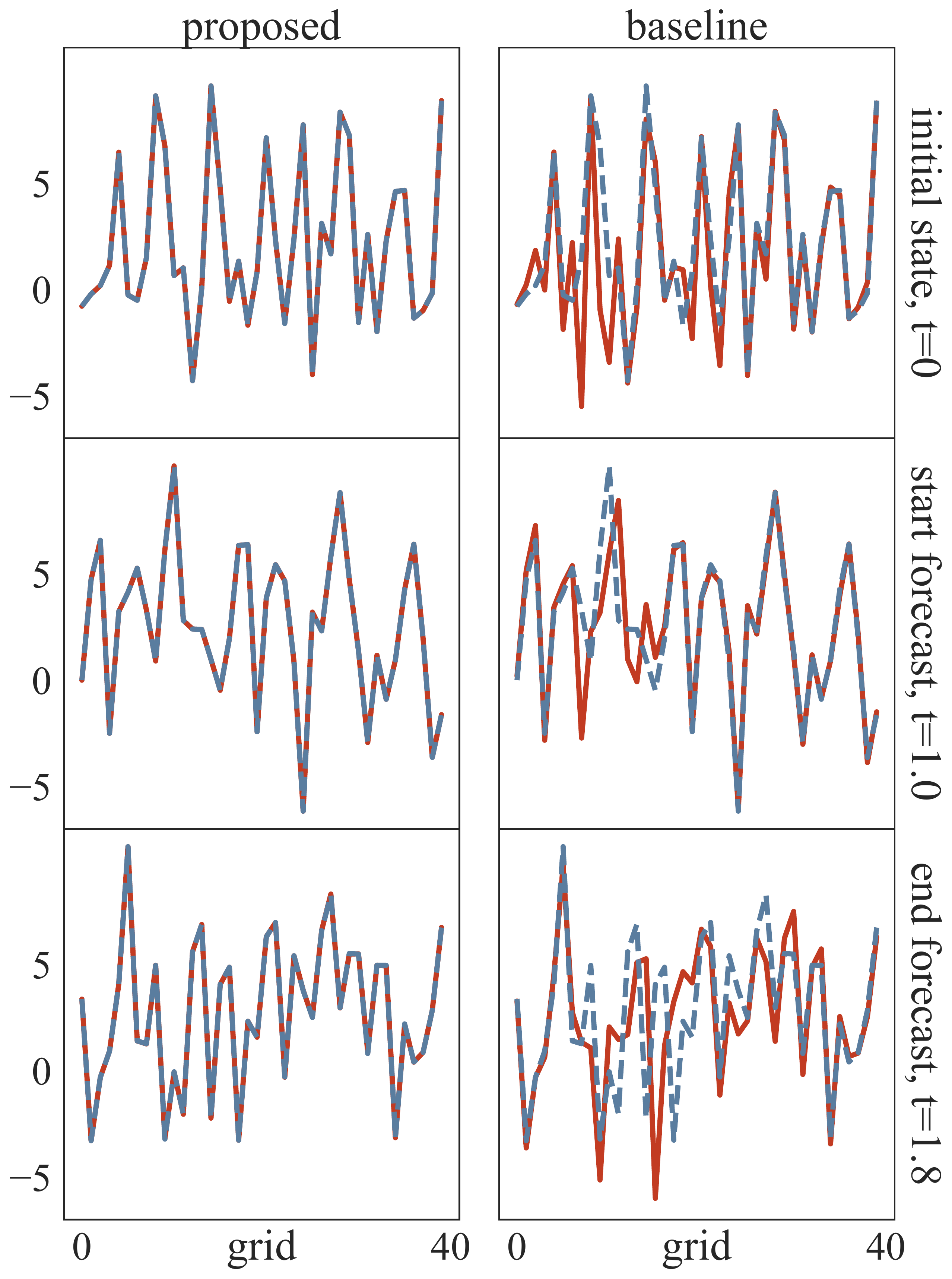}
    \caption{Forecast trajectory for the Lorenz96 model optimized from the initial conditions depicted in Figure~\ref{fig:da_init_lorenz96}. The forecast trajectory based on optimization with the inverse observation operator (inverse initialization, hybrid optimization) is qualitatively much closer to the ground truth evolution of the system than the baseline method (averaging initialization, observation space optimization).}
    \label{fig:forecast_lorenz96}
\end{figure}

Figure~\ref{fig:forecast_lorenz96} shows an example forecast of the system. 
The hybrid method initialized with the learned inverse mapping is able to capture the ground truth evolution of the system.
In contrast, for the baseline method a visible approximation error remains throughout the system integration.

\subsection{Two-Dimensional Turbulence}
\label{sec:kolmogorov_flow}
Next, we study data assimilation for a two-dimensional turbulent fluid \cite{Boffetta2012-gp}.
Machine learning in this setting requires modeling richer physics and poses a much more demanding computational problem.
Furthermore, having in mind the application of data assimilation to numerical weather prediction, this class of models can be considered as the simplest approximation to modeling the flow of the atmosphere.

\begin{figure}[t!]
    \centering
    \includegraphics[width=0.7\columnwidth]{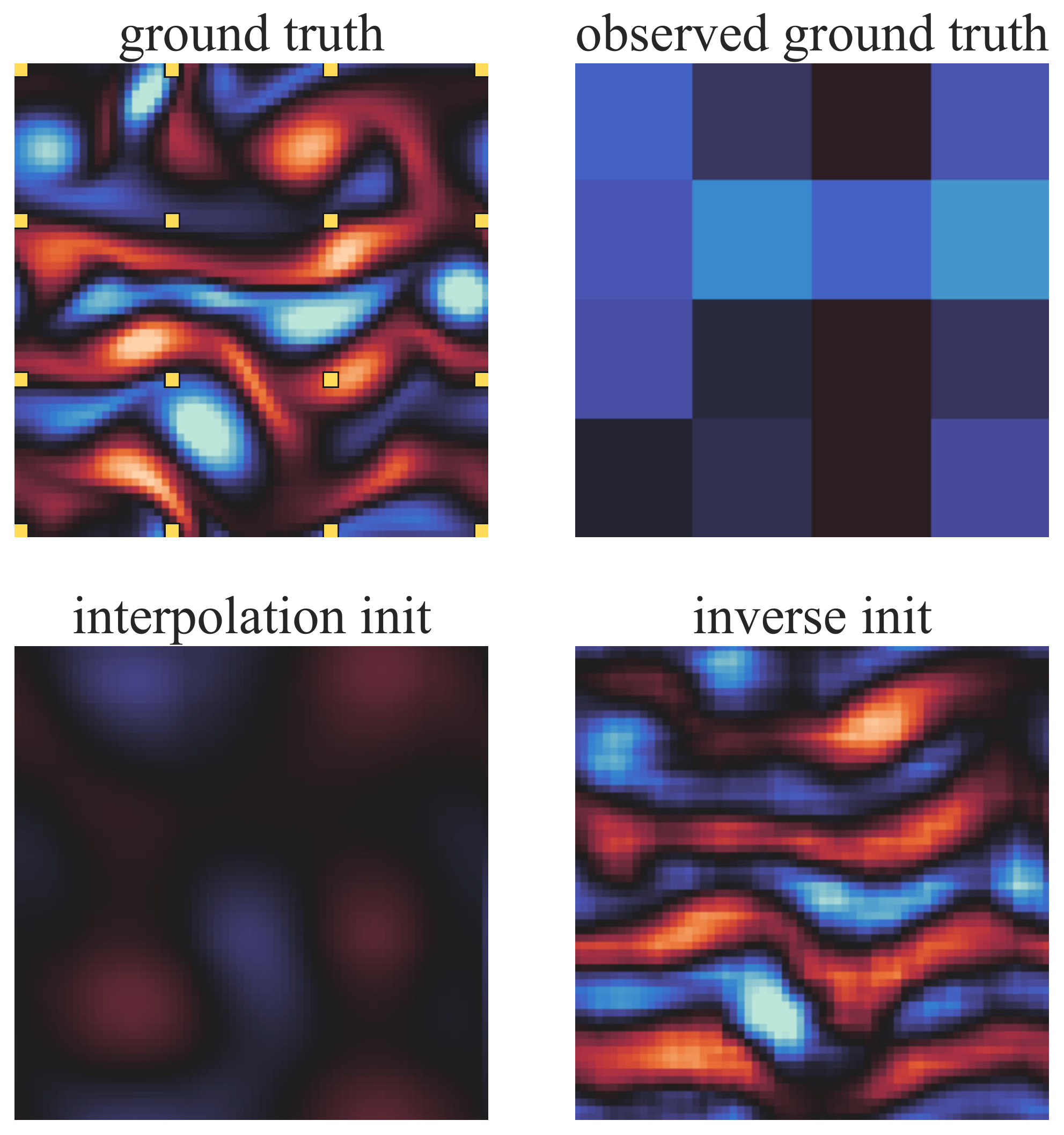}
    \caption{
        Comparison of initialization schemes with the ground truth initial state for Kolmogorov flow with a $16$-subsampling observation operator (yellow dots). 
        The learned inverse observation operator is trained on a trajectory of subsampled velocity fields.
        The observed ground truth vorticity exemplifies the amount of information of a single state of this trajectory.
        The trained model predicts a good approximation to the ground truth state only from a sequence of $4\times 4$ points.
    }
    \label{fig:da_init_kolmogorov}
\end{figure}

\begin{figure}[t!]
    \centering
    \includegraphics[width=\columnwidth]{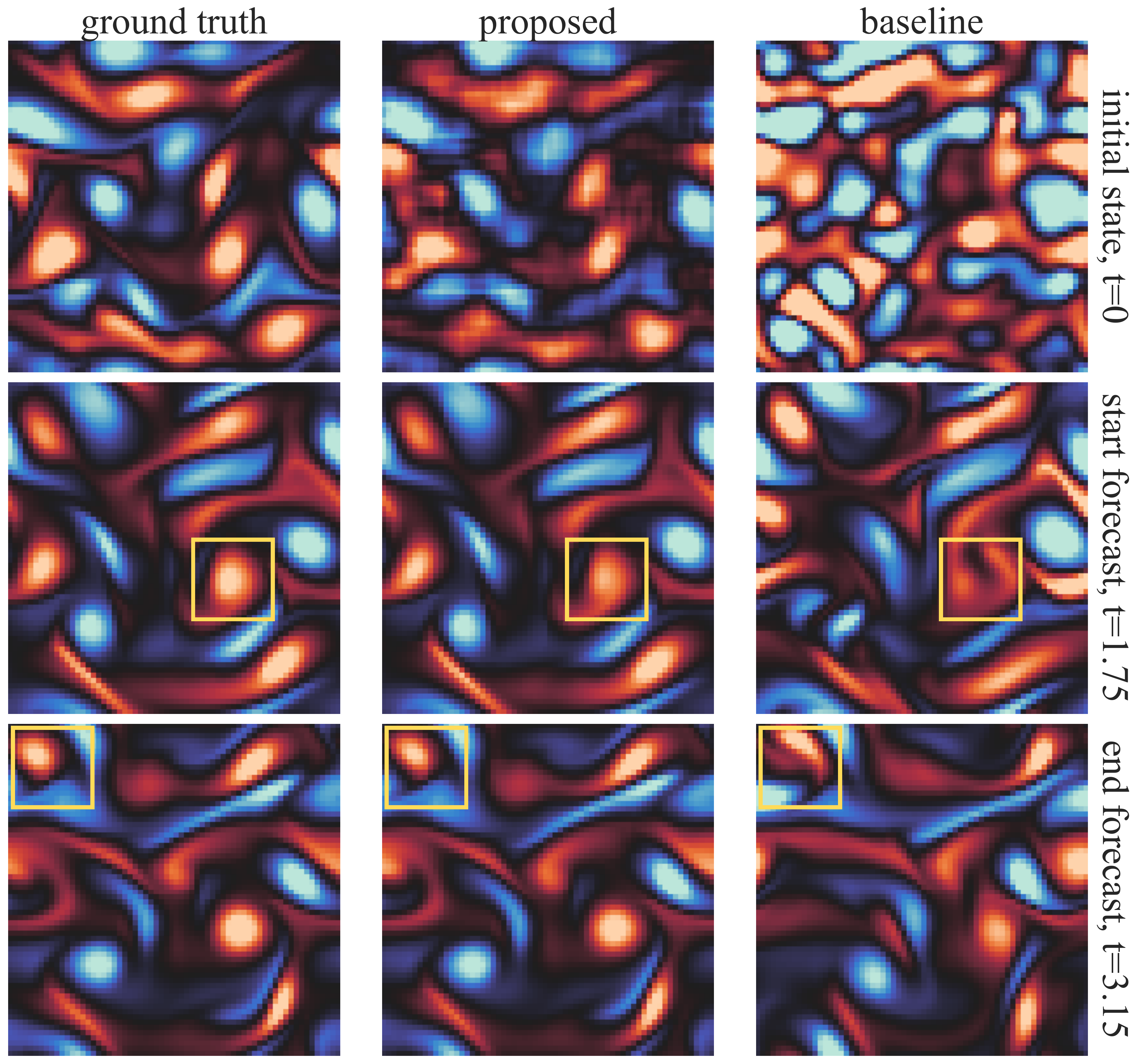}
    \caption{
        Vorticity snapshots of a forecast trajectory for Kolmogorov flow.
        The proposed method (inverse initialization, hybrid optimization) more accurately captures ground truth vorticity features (yellow squares) compared with the baseline (interpolation initialization, observation space optimization).
        Depicted are the initial state and snapshots from the start and end of the forecast window.
        Note that certain perturbations vanish quickly during the evolution of the system and are therefore not optimized to vanish in the initial state.
    }
    \label{fig:forecast_kolmogorov}
\end{figure}

We consider the incompressible Navier-Stokes equation for a velocity field $\mathbf{u}$ and a pressure field $p$:
\begin{align}
    \frac{\partial \mathbf{u}}{\partial t} + \mathbf{u}\nabla \mathbf{u} - \nu \nabla^2 \mathbf{u} + \nabla p - \mathbf{F} &= 0 \\
    \nabla \cdot \mathbf{u} &= 0 \nonumber \;,
\end{align}
where $\nu$ is the kinematic viscosity of the fluid.
We choose the external forcing $\mathbf{F}$ to correspond to Kolmogorov flow \cite{Chandler2013}, with linear damping \cite{Boffetta2012-gp} to ensure that the long-time behavior of the solution is statistically stationary:
\begin{align}
    \mathbf{F} = \sin(kx) \mathbf{\hat x} - \alpha \mathbf{u}
\end{align}
For our experiments, we choose a domain $[0, 2\pi]^2$ with periodic boundary conditions, a wavenumber $k=4$, a damping coefficient $\alpha=0.1$, and 
a viscosity of $\nu=10^{-2}$.
We discretize the solution on a $64 \times 64$ grid and use standard numerical methods to solve the Navier-Stokes equation with a differentiable solver written in JAX \cite{Kochkov2021-sh}.
The Lyapunov time of the system is approximately $5.9$ time units.
Our flows are initialized with a random velocity field filtered with a spectral filter at a peak wavenumber $4$, which is then integrated to a statistically stationary regime of the flow.
We assimilate over trajectories of length $T=10$ time steps, where the integration time between two such snapshots is  $\Delta t\approx 0.18$, consisting of 25 internal solver integration steps.
To save memory when computing gradients, we checkpoint the state from the forward pass only at each internal integration step rather than storing intermediate values~\cite{Griewank1994-yg}.
This requires evaluating the forward pass twice, but reduces memory usage by two orders of magnitude.\\
We carry out data assimilation on the velocity field $\mathbf{u}$ of the flow.
To analyze our forecasts, we use vorticity $\omega$, the curl of the velocity field,
\begin{align}
    \omega \coloneqq \left(\frac{\partial \mathbf{u}_y}{\partial x} - \frac{\partial \mathbf{u}_x}{\partial y}\right) \;.
\end{align}
Vorticity describes the local direction of movement of the fluid.
We visualize vorticity on a scale, which is cut off at $[-8, 8]$ with negative values (blue in Figures~\ref{fig:da_init_kolmogorov}~and~\ref{fig:forecast_kolmogorov}) denoting clockwise rotation and positive values (red in Figures~\ref{fig:da_init_kolmogorov}~and~\ref{fig:forecast_kolmogorov}) denoting counter-clockwise rotation.
\begin{table}[t!]
    \caption{
        Fully-convolutional network for training the inverse observation operator for Kolmogorov flow. 
        The table shows a layer with its respective output array dimensions time (\textsc{T}), space (\textsc{X} and \textsc{Y}), and channel (\textsc{C}).
        The \textsc{Conv3D} layer applies periodic convolution in the two space dimensions and zero-padded convolution in the time dimension.
        The filter size for all convolutional layers is $(3,3,3)$.
        \textsc{BN} denotes batch normalization.
        We upsample the grid using bicubic interpolation by a factor of $2$ and correspondingly halve the number of channels.
        As a non-linearity we use the sigmoid-weighted linear unit (\textsc{SiLU}), $\mathrm{silu}(x) = x / (1 + \exp(-x))$.
    }
    \vskip 0.15in
    \centering
    \begin{small}
    \begin{sc}
    \begin{tabular}{ll}
        \toprule
         Layer &  (T, X, Y, C) \\
         \midrule
         Input & (10, 4, 4, 2) \\
         Conv3D + BN + SiLU & (10, 4, 4, 64) \\
         Upsample + Conv3D + BN + SiLU & (10, 8, 8, 32) \\
         Upsample + Conv3D + BN + SiLU & (10, 16, 16, 16) \\
         Upsample + Conv3D + BN + SiLU & (10, 32, 32, 8) \\
         Upsample + Conv3D + BN + SiLU & (10, 64, 64, 4) \\
         Conv3D  & (10, 64, 64, 2) \\
         \bottomrule
    \end{tabular}
    \end{sc}
    \end{small}
    \label{tab:kolmogorov_net}
\vskip -0.1in
\end{table}\\
We again use  equidistant subsampling for the observation operator $\H$. In contrast to the Lorenz96 model the solution is smooth over the grid points, so we can use bicubic interpolation between  observed grid points as the baseline initialization scheme. 
We now demonstrate the effect of a learned inverse observation operator, when this operator observes every $16$th grid point.
For training, we employ a fully-convolutional network as shown in Table~\ref{tab:kolmogorov_net}.
We train on a dataset of $32000$ independent observation trajectories with batch size $8$ for $500$ epochs.

For data assimilation, we compare bicubic interpolation as the baseline \textit{interpolation initialization} with \textit{inverse initialization} derived from the learned inverse observation operator, as depicted in Figure~\ref{fig:pipeline_sketch}.
Figure~\ref{fig:da_init_kolmogorov} compares these initialization methods with the ground truth initial state.
As with the Lorenz96 model analyzed in Section~\ref{sec:lorenz96}, we also compare optimizing in observation space with the hybrid approach of first optimizing in physics space and subsequently refining this solution in optimization space.
We limit both optimization methods to $1000$ steps, with the hybrid approach using $100$ of these steps to optimize in physics space and the remaining $900$ to refine in observation space.
\begin{figure}[tbh!]
    \centering
    \includegraphics[width=0.95\columnwidth]{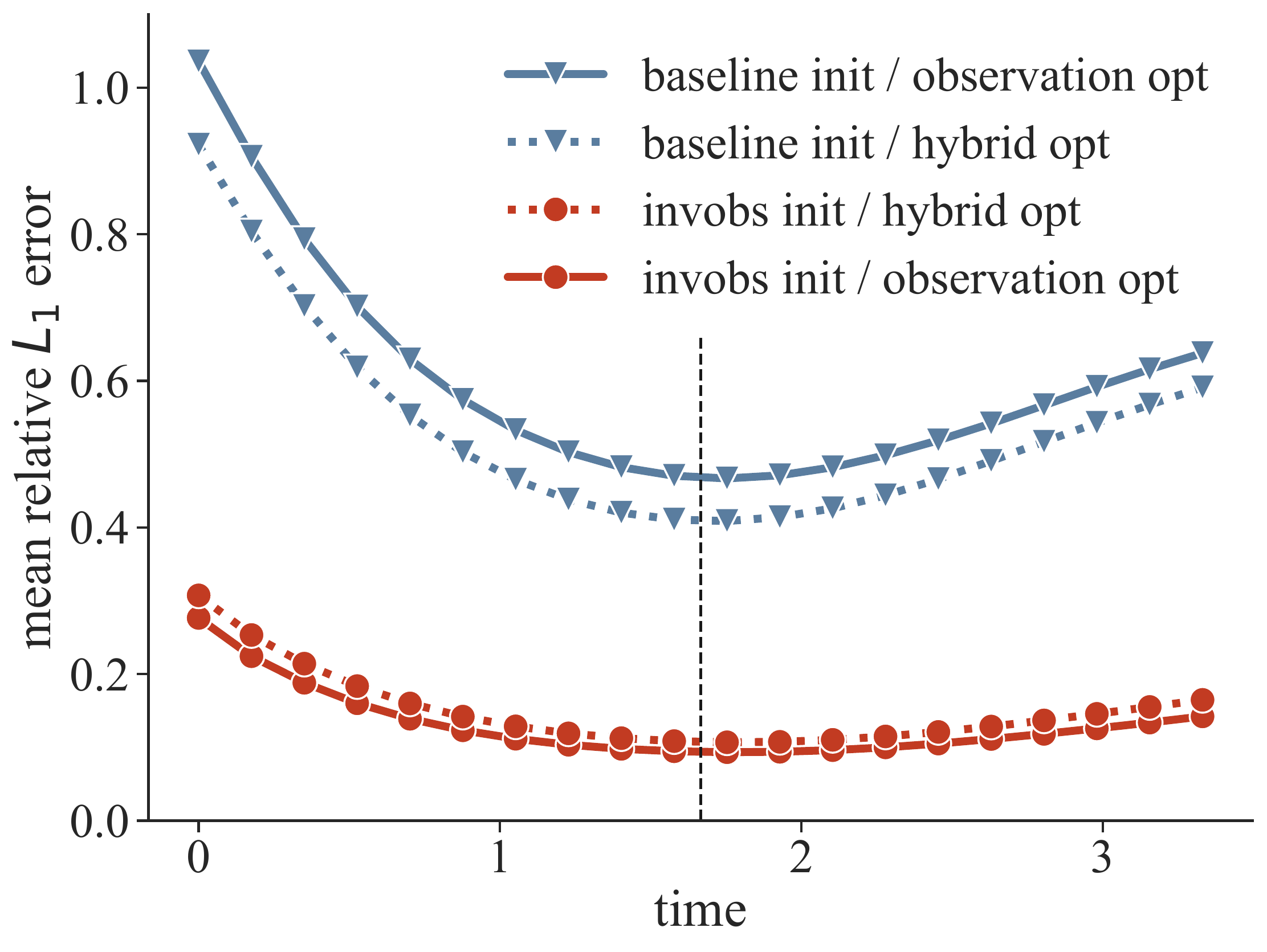}
    \caption{
        Data assimilation results for Kolmogorov flow using a learned inverse observation operator. 
        Quality measure is the $L_1$ forecast error relative to a random evolution of the system. 
        Depicted is the mean error based on a sample of $100$ trajectories.
        Trajectories are obtained by evolving the initial states returned by corresponding optimization methods.
        Using the hybrid method for optimization improves assimilation quality with inaccurate interpolation initialization.
        The inverse initialization scheme significantly improves forecasts for observation space optimization.
        Adding inverse initialization to hybrid optimization does not improve performance.
        The difference between observation space optimization vs. hybrid optimization for both initialization schemes is significant with a p-value of $p < 10^{-8}$.
    }
    \label{fig:da_kolmogorov_invobs}
\end{figure}
Figure~\ref{fig:da_kolmogorov_invobs} shows the results, with three implications analogous to experiments on the Lorenz96 model.
First, hybrid optimization improves assimilation quality even when using the inaccurate interpolation initialization.
This implies that by first optimizing in physics space, we arrive at a favorable basin of attraction for optimization in observation space.
Secondly, using inverse initialization for optimizing in observation space significantly improves forecasts.
Finally, adding inverse initialization to hybrid optimization does not improve performance. 
This is presumably because the initialized state is already a good approximation to the optimization target, so there is no added advantage in optimizing in physics space.
In contrast, it is more sensible to directly refine this state by optimizing in observation space.
\begin{figure}[t!]
    \centering
    \includegraphics[width=\columnwidth]{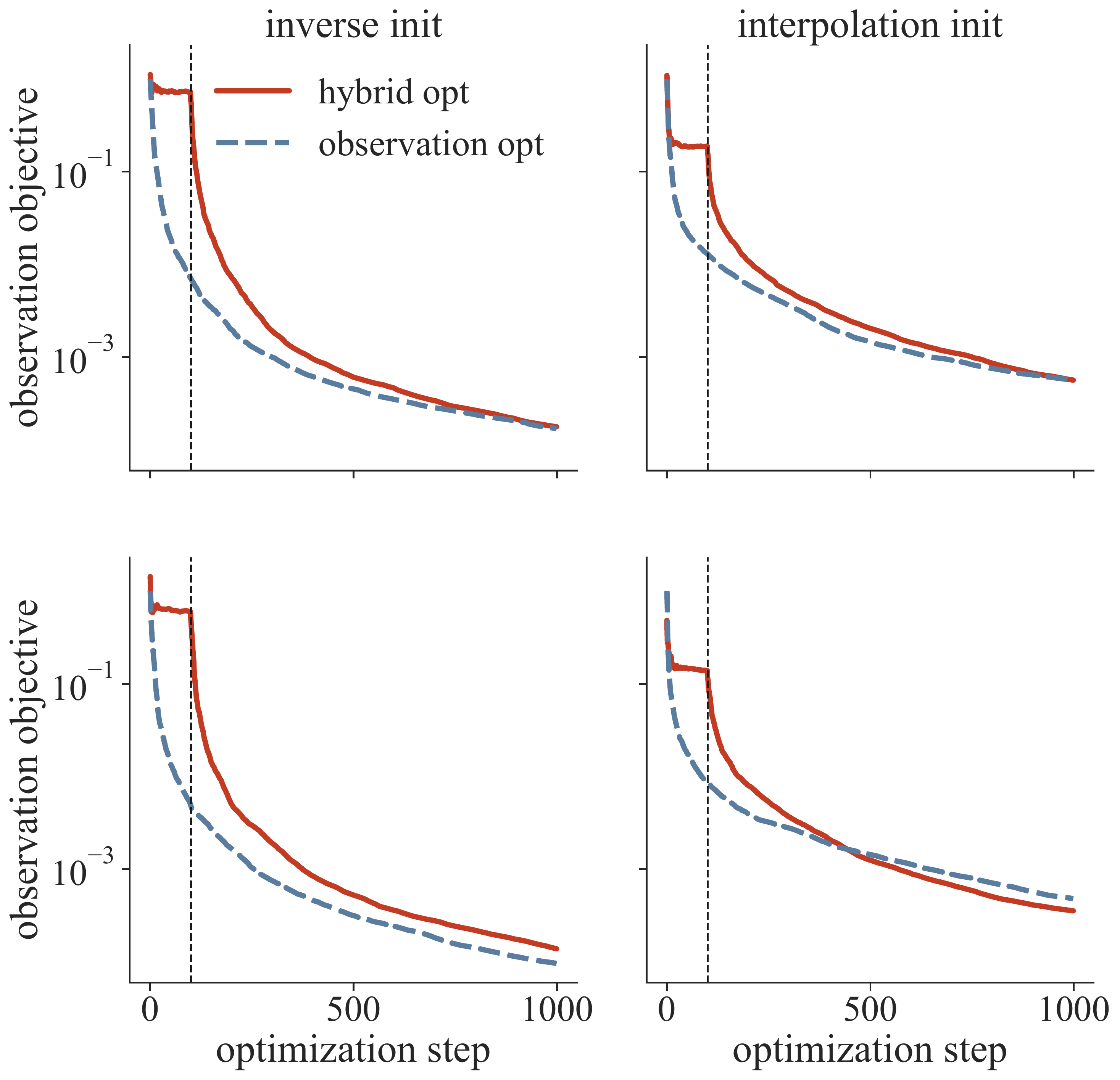}
    \caption{
        Observation space data assimilation objective \eqref{eq:da_obs_space} during optimization.
        Values are relative to the initial value on a log-scale for two different samples (rows).
        Depicted are observation space optimization (dashed blue) and hybrid optimization (solid red).
        For both methods, we evaluate the \textit{same} observation space objective function along the optimization path.
        The vertical dashed line signifies the change from physics space to observation space in the hybrid method. 
        For inaccurate interpolation initialization, a favorable basin of attraction can be reached for some samples by first optimizing in physics space.
        Inverse initialization provides such a good initial condition that there is no added advantage of first optimizing in physics space.
    }
    \label{fig:opt_curves_kolmogorov}
\end{figure}
This effect becomes evident when analyzing how each of these optimization methods decreases the objective function in observation space during optimization, as depicted in Figure~\ref{fig:opt_curves_kolmogorov}.
For the less accurate interpolation initialization, a more favorable basin of attraction can be reached for some samples by first optimizing in physics space.
However, since inverting the observation space trajectory only approximates the true trajectory, fitting against this target precludes progress after an initial phase of optimization steps.
Hence, with a fixed budget of optimization steps there is a trade-off between finding a favorable basin of attraction by optimizing in physics space and finding a higher-accuracy solution by optimizing in observation space.
Figure~\ref{fig:forecast_kolmogorov} qualitatively compares vorticity forecasts for the baseline method (interpolation initialization, observation space optimization) with our method based on the inverse observation operator for initialization and hybrid space optimization.
The dominant features of the flow are visibly better predicted by the proposed method.
Note that the structure of initial states differs from that of the following trajectories.
Certain perturbations vanish quickly during the evolution of the system and are therefore not optimized to vanish in the initial state.

\subsection{Result Summary}
\label{sec:result_summary}
To summarize the results, we compare the relative performance of each optimization setting for the first forecast state, as depicted in Table~\ref{tab:result_summary}.
By using the learned inverse observation operator, the forecast error can be significantly reduced for both models.
The relative merit of exploiting this operator for initialization and transformation of the objective function depends on the properties of the physical model.
For the Lorenz96 model, hybrid optimization in addition to inverse initialization notably improves performance. 
For Kolmogorov flow, the learned inverse mapping already provides an extremely good initialization and hence optimizing in physics space does not further reduce the forecast error.
\begin{table}[t!]
    \caption{
        Mean $L_1$ forecast error of the first forecast state.
        All values are relative to the baseline method for the respective model.
        The table compares both initialization schemes (baseline, inverse) and optimization methods (observation space, hybrid) for the Lorenz96 model and Kolmogorov flow.
        The best optimization setting is emphasized in bold face.
        Using the learned inverse observation operator improves optimizability for both models.
    }
    \vskip 0.11in
    \centering
    \begin{small}
    \begin{sc}
    \begin{tabular}{l|cc|cc}
        & \multicolumn{2}{c|}{Lorenz96} & \multicolumn{2}{c}{Kolmogorov} \\
        & obs & hybrid & obs & hybrid \\
        \midrule
         baseline & 1 & 0.08 & 1 & 0.88 \\
        inverse & 0.25 & \textbf{0.07} & \textbf{0.20} & 0.23 \\
        \bottomrule
    \end{tabular}
    \end{sc}
    \end{small}
    \label{tab:result_summary}
\vskip -0.1in
\end{table}

\section{Conclusion}
Data assimilation is the perfect problem class to explore the combination of physical modeling and machine learning since applications naturally involve rich physics and vast amounts of data.
We demonstrate in this paper that a traditional variational data assimilation pipeline is improved by using a learned inverse observation operator. 
Exploiting this operator, we transform the 4D-Var optimization problem and show significantly enhanced forecast quality on two canonical chaotic models, the Lorenz96 model and a two-dimensional turbulent fluid flow.
More broadly, our work shows that the core functionality of modern machine learning frameworks -- support for automatic differentiation, hardware accelerators and deep learning -- can advance research for data assimilation and other physics constrained optimization problems.

\clearpage
\section*{Acknowledgement}
We thank Stephan Rasp, Casper S{\o}nderby, and Jason Hickey for fruitful discussions on the topic.
This work was conducted during an internship of T.F. at Google Research and was supported by the Munich Center for Machine Learning.
\bibliography{neural_da}
\bibliographystyle{icml2021}
\end{document}